\documentclass[conference]{IEEEtran}
\IEEEoverridecommandlockouts

\usepackage{cite}
\usepackage{amsmath,amssymb,amsfonts}
\usepackage{algorithmic}
\usepackage{graphicx}
\usepackage{textcomp}
\usepackage{xcolor}
\usepackage{float}
\usepackage{url}
\usepackage{subcaption}
\usepackage{placeins}
\usepackage{capt-of} 

\def\BibTeX{{\rm B\kern-.05em{\sc i\kern-.025em b}\kern-.08em
    T\kern-.1667em\lower.7ex\hbox{E}\kern-.125emX}}

\newcommand{\scumthreec}{SC$\mu$M-3C}

\begin{document}

\title{Control of Microrobots with Reinforcement Learning under On-Device Compute Constraints
}

\author{
\IEEEauthorblockN{Yichen Liu\IEEEauthorrefmark{1}, Kesava Viswanadha\IEEEauthorrefmark{1}, Zhongyu Li, Nelson Lojo, Kristofer S. J. Pister}
Electrical Engineering and Computer Science Dept., University of California, Berkeley, USA \\
Email: lyichen@eecs.berkeley.edu
}

\maketitle
\def\thefootnote{*}\footnotetext{\textbf{Both authors contributed equally to this paper.}}\def\thefootnote{\arabic{footnote}}

\begin{abstract} An important function of autonomous microrobots is the ability to perform robust movement over terrain. This paper explores an edge ML approach to microrobot locomotion, allowing for on-device, lower latency control under compute, memory, and power constraints. This paper explores the locomotion of a sub-centimeter quadrupedal microrobot  via reinforcement learning (RL) and deploys the resulting controller on an ultra-small system-on-chip (SoC), SC$\mu$M-3C, featuring an ARM Cortex-M0 microcontroller running at 5 MHz. We train a compact FP32 multilayer perceptron (MLP) policy with two hidden layers ($[128, 64]$) in a massively parallel GPU simulation and enhance robustness by utilizing domain randomization over simulation parameters. We then study integer (Int8) quantization (per-tensor and per-feature) to allow for higher inference update rates on our resource-limited hardware, and we connect hardware power budgets to achievable update frequency via a cycles-per-update model for inference on our Cortex-M0. We propose a resource-aware gait scheduling viewpoint: given a device power budget, we can select the gait mode (trot/intermediate/gallop) that maximizes expected RL reward at a corresponding feasible update frequency. Finally, we deploy our MLP policy on a real-world large-scale robot on uneven terrain, qualitatively noting
that domain-randomized training can improve out-of-distribution
stability. We do \textit{not} claim real-world large-robot empirical zero-
shot transfer in this work.
\end{abstract}


\section{Introduction}

Intelligent crawling microrobots based on biomimetic mechanisms have received increasing attention in recent years. Due to their small size and versatility, microrobots have been proposed for applications ranging from planetary exploration \cite{b2} to minimally invasive surgery \cite{b4}. While significant progress has been made in hardware development, including actuation, fabrication, and power efficiency, reliable locomotion control for sub-centimeter terrestrial microrobots remains challenging, particularly on complex terrain. Physical experimentation is constrained by the difficulty of constructing representative test environments and the high cost of iterative testing. Therefore, simulation-based approaches provide a scalable and cost-effective framework for accelerating controller development and experimental iteration.

A major challenge regarding microrobot control is the on-device autonomy. Sophisticated control algorithms for larger walking robots can be implemented on high-performance CPUs with lower concern for power constraints \cite{b36}. In contrast, the control algorithm for micro-scale walking robots is limited by the size, computing performance, and power efficiency of the controller hardware. Microrobots must often rely on off-board computation or open-loop gait patterns triggered by external stimulation. Recent work has moved towards microrobot autonomy by performing computation on-device (e.g., on-board digital control for microscopic and millimeter systems \cite{b25,b26,b27,b28}), but on-device, low-latency, RL-based locomotion for terrestrial microrobots remains under-explored relative to macro-scale quadrupeds.

\begin{figure}[htbp]
\centering
\includegraphics[width=0.47\textwidth]{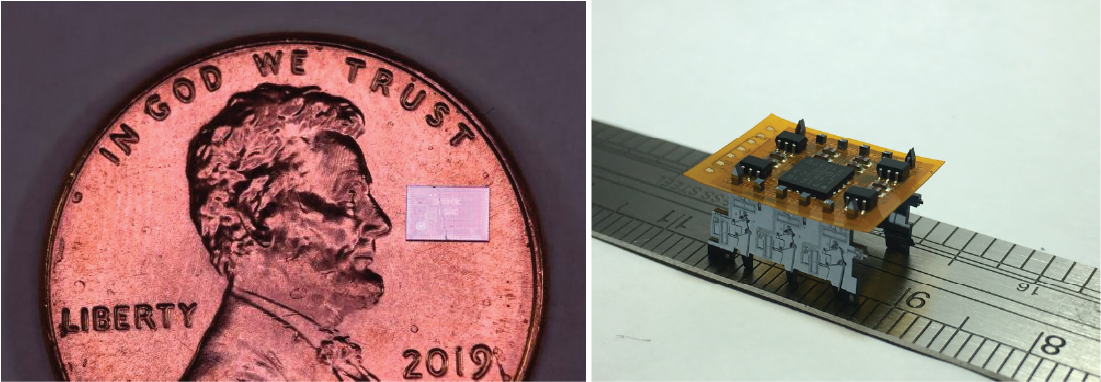}
\caption{Single-Chip Micro Mote, SC$\mu$M (left) and SOI-based microrobots (right).}
\label{fig:Intro1}
\end{figure}

Deploying a gait controller approach on a micro-scale robot is restrictive, aside from the challenges of training and optimizing the controller. This has typically driven researchers towards off-board processing \cite{b6} \cite{b7} or embedding gait in hardware that is activated using external stimulation \cite{b8}. Yet, on-board control remains more responsive and versatile. A truly autonomous micro-robot will be able to maneuver across complex terrain independently without the restriction of a wireless connection. Our group has designed the single-chip micro mote, SC$\mu$M, a compact microrobot control system as shown in Figure \ref{fig:Intro1} \cite{b11}. Our group has also analyzed a central pattern generator (CPG)-based algorithm for micro-scale crawlers, which has shown promising results in simulation \cite{b5}.

Reinforcement learning (RL) has demonstrated strong results for robust locomotion and Sim2Real transfer on quadrupeds \cite{b12} \cite{b13}. This paper explores RL-based locomotion for sub-centimeter quadrupedal microrobots trained in simulation and executed under strict on-device constraints on the SC$\mu$M microchip \cite{b9} \cite{b11}. We focus not on whether microrobots can perform sufficient on-board computation (evidenced by \cite{b25,b26, b27, b28}), but on connecting the learned locomotion reward and policy robustness, the update-frequency dependent reward of gaits, and device-level power and computation constraints into a gait-selection methodology that supports deployment decisions.

Our paper focuses on the following:
\begin{enumerate}
    \item Training an MLP policy for sub-centimeter quadrupedal microrobot locomotion in massively parallel GPU simulation, improving robustness via domain randomization.
    \item Analyzing reward trends under changes in inference update frequency and quantization while validating on-device feasibility with \scumthreec.
    \item Deriving cycle and power models for Int8 inference on a Cortex-M0-class core and proposing a resource-aware gait-selection method that selects the gait maximizing expected reward under given hardware constraints.
\end{enumerate}

\section{Related Work}

\subsection{Microscale Terrestrial Walkers and Controller Design}
Walking silicon microrobots and MEMS-based terrestrial microrobots have a long history (e.g. early silicon walking microrobots \cite{b14}). 
In many insect-scale legged microrobots, locomotion has often been demonstrated in tethered settings using off-device power and control electronics, but high-fidelity wireless locomotion has also recently been demonstrated \cite{b29}. Our group has previously demonstrated the use of multi-objective optimization of the locomotion controller
on a  hexapod microrobot \cite{b5}. However, the
presented works either don't use fully on-device control or utilize open-loop controls, making the system vulnerable to interference and noise in practice.

\subsection{Learning-Based Control for Microrobot Locomotion}
Learning-based gait design has been studied for microrobot-inspired platforms \cite{b5} \cite{b30} \cite{b31}. Yang \emph{et al.} \cite{b5} formulate microrobot locomotion as a data-efficient controller-search problem, using contextual optimization to learn reusable locomotion primitives for a silicon microrobot and then using these primitives for higher-level behaviors like maze navigation. 
Complementary to this work, Lambert \emph{et al.} \cite{b31} propose avenues for improving model-based RL control for microrobots by clustering and filtering out redundant data, improving data and power efficiency. Lambert \emph{et al.} also propose MicroBotNet, a $<10^6$ MAC network that provides visual Simultaneous Localization and Mapping (SLAM) for navigation. Together, these works motivate learning-based locomotion at microrobot scales while highlighting hardware constraints as a primary design consideration.

\subsection{Sim2Real and Domain Randomization for Legged Locomotion}

Simulation-based optimization has proven to be effective in generating training data for locomotion control on macro-scale legged robots. The degradation in solution performance between simulation and reality has also been well-characterized \cite{b32}. Despite the success in macro-scale Sim2Real control, it is difficult to accurately adapt simulation to reality on the sub-centimeter scale. At these length scales, the error in empirical model approximations of quantities like dry friction is amplified due to the complicated mechanics of contact at such a small scale. Physically accurate simulation at this scale is very computationally expensive. To address the issues of error and noise, researchers often resort to analyzing and simulating simplified robot configurations or using scaled-up models to enhance robustness. These approaches are typically not adequate for more intricate small-scale mechanisms, as this oversimplification creates disparity between simulation and real-world performance. We consider the technique of domain randomization \cite{b16} to overcome the reality gap challenge in an RL-based robot controller. While domain randomization for the locomotion of HAMR, a 4.5 cm long, 1.43 g quadruped, has been explored \cite{b33}, to our knowledge domain randomization for sub-centimeter scale microrobot locomotion has not been well-explored. 

\subsection{On-Board Control and Computation at Micro/Milli Scales}
Recent work has pushed autonomy at very small scales. Bandari \emph{et al.} present $\le$1\,mm edge length ambient light power-harvesting microrobots that include an on-board microcontroller and sensors, and show locomotion steered by sensory feedback from environmental stimuli\cite{b28}. This work establishes meaningful autonomy at milli scales, but does not focus on RL-trained terrestrial locomotion policies.

\subsection{Edge ML / TinyML and Quantized Inference}
TinyML deployments commonly use Int8 quantization and optimized kernels (e.g. CMSIS-NN) to execute neural networks efficiently on Cortex-M class microcontrollers \cite{b18}. With improved space efficiency due to quantization, we can have on-device model storage. Important advantages of local computation are reliability and latency reduction. Robot locomotion requires high update frequency between sensing and actuation, especially across complex terrain, as latency and packet loss would result in poor locomotive performance. 

\section{Methodology}
The objective of our experiment is two-fold: (i) optimize a locomotion controller for a sub-centimeter terrestrial microrobot in simulation using reinforcement learning, and (ii) evaluate deployment feasibility under on-device compute constraints by running the learned policy on-device in a higher-fidelity closed-loop model.

\subsection{Physical Model and Robot Configuration}
\begin{figure}[htbp]
\centering
\includegraphics[width=0.47\textwidth]{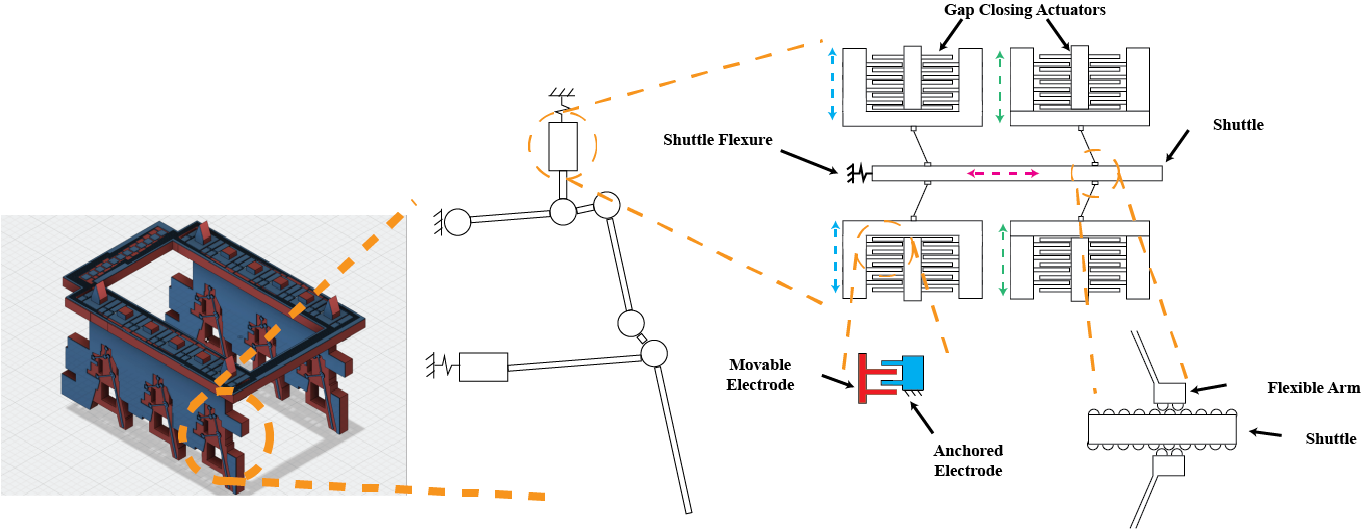}
\caption{SOI MEMS Hexapod Walker: legs, inchworm motor, and actuator configuration.}
\label{fig:Method_hw}
\end{figure}

The physical configuration of the quadruped microrobot is adapted from a hexapod walker architecture fabricated in an SOI MEMS process described in \cite{b34}. Each leg is composed of electrostatically driven gap-closing actuators (GCAs), which move a shuttle back and forth to form linear inchworm motors. These in turn actuate a pin-and-joint leg linkage (Fig.~\ref{fig:Method_hw}). Each leg is driven by two orthogonal prismatic actuators (``$x$'' and ``$y$'' motors) that produce in-plane motion with two degrees of freedom (DoFs).

The above robot system is constructed from three joined chips: two vertical chips housing actuators and legs mechanically and electrically connected to a third chip for computation, energy harvesting, and power storage. A representative fabricated hexapod has dimensions of approximately $13~\mathrm{mm}\times 9.6~\mathrm{mm}\times 7~\mathrm{mm}$ and mass $\sim 200~\mathrm{mg}$. A quadrupedal version is under development with similar specifications and actuation primitives.

\begin{figure}[htbp]
\centering
\includegraphics[width=0.5\textwidth]{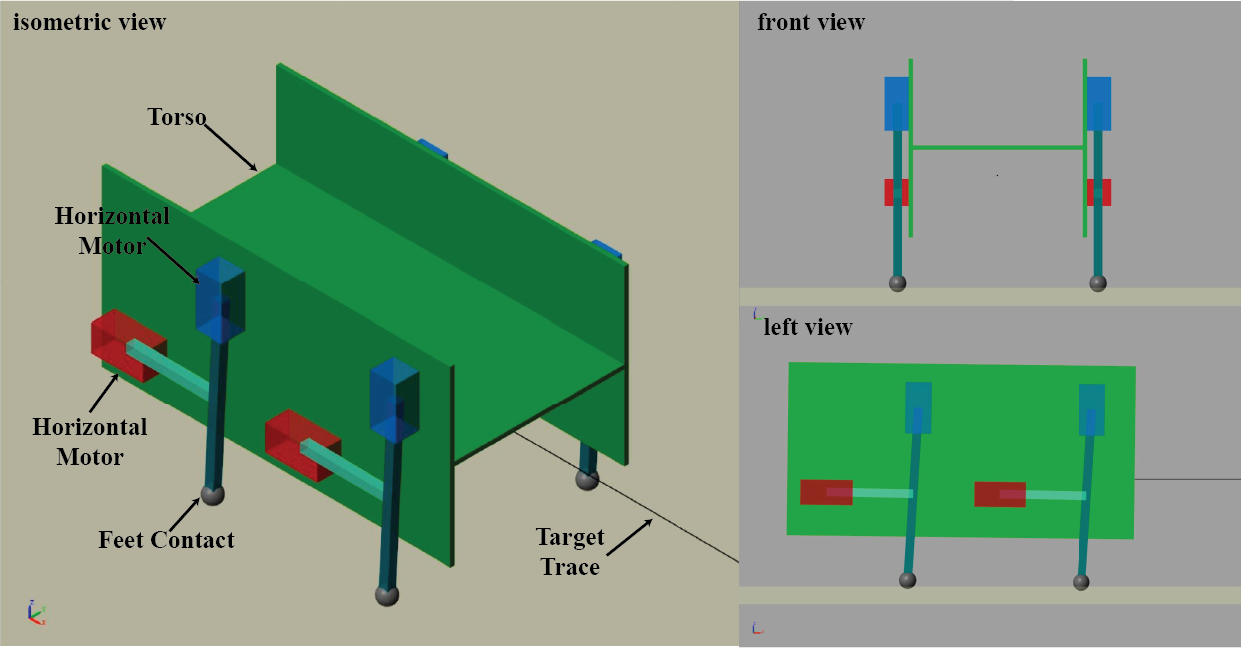}
\caption{Quadruped configuration in Simulink Multibody.}
\label{fig:Method_matlab}
\end{figure}

Designing dynamic models for the legged robot presents unique challenges due to the prismatic motors and mechanical linkages. The prismatic motors require consideration of interfaces between electrostatic forces and mechanical interactions, while the mechanical linkages have complex rotational friction effects that are not well studied. To capture the intended dynamics at evaluation time, we use a Simulink Multibody model of an 'H'-shaped torso with two legs on each side (Fig.~\ref{fig:Method_matlab}). Each leg is modeled with two prismatic joints with positional inputs and two linkages connected by rotary joints in a closed-loop configuration, with a joint/actuator model based on electrostatic forces between inchworm motors \cite{b35}. This configuration is used for evaluation/inferencing whereas an open-loop configuration is used for training.

\subsection{Network Training and Structure (PhysX / Isaac Gym)}
We train the locomotion controller using Nvidia Isaac Gym \cite{b20} with PhysX simulation and proximal policy optimization \cite{b22}. Isaac Gym enables massively parallel rollout collection as well as GPU-accelerated policy updates. This training style is commonly used to increase simulation throughput for legged locomotion \cite{b12}. In our setup, 4096 agents are trained in parallel at $f_{\text{train}}=120~\mathrm{Hz}$ with episode length $10~\mathrm{s}$. We use a rollout length of 64 steps per update (approximately $0.5~\mathrm{s}$ of simulated time).

There are two input commands for the agents: forward speed along the torso and angular velocity for turning. Observations include torso linear and angular velocity, the gravity vector expressed with respect to the base, joint positions and velocities, and the previous action selected by the policy. All of these described quantities compose the input to our policy. No additional sensor information is provided to compensate for the limited sensing on our microrobots.

The total reward is a sum of five terms listed in Table~\ref{tab:reward_function_terms}. The reward parameters themselves are listed in Table~\ref{tab:reward_parameters}. Two rewards are given for following the provided linear and angular velocity commands. An additional airtime reward is given to encourage natural motion by measuring the time between instances of microrobot feet contacting the ground. Lastly, two punishments are given to undesirable linear and angular velocity on the base to discourage jitter and other unnatural motion patterns.

\begin{table}[H]
\centering
\caption{Reward parameter symbols.}
\label{tab:reward_parameters}
\begin{tabular}{ll}
\hline
\textbf{Parameter Names}     & \textbf{Symbols} \\\hline
Joint positions              & $q_{j}$          \\
Joint velocity               & $\dot{q}_{j}$    \\
Target joint positions       & $q_{j}^{*}$      \\
Base linear velocities       & $v_{b}$          \\
Base angular velocity        & $\omega_{b}$     \\
Base pitch and roll          & $a_{b}$          \\
Commanded base velocity      & $v_{b}^{*}$      \\
Commanded angular velocities & $\omega_{b}^{*}$ \\
Environment time step        & $dt$             \\\hline
\end{tabular}
\end{table}

\begin{table}[H]
\centering
\caption{Reward terms and weights used in training.}
\label{tab:reward_function_terms}
\begin{tabular}{lll}
\hline
Parameters                & Definition                             & Weight   \\\hline
Linear velocity tracking  & $\Phi(v_{b,x}^{*}-v_{b,x})$            & $1 \cdot dt$    \\
Angular velocity tracking & $\Phi(\omega_{b,z}^{*}-\omega_{b,z})$  & $0.5 \cdot dt$  \\
Linear velocity penalty   & $- v_{b,y}^{2}$       & $0.5 \cdot dt$  \\
Angular velocity penalty  & $- \omega_{b,xy}^{2}$ & $0.05 \cdot dt$ \\
Feet air time             & $\sum_{f}(t_{air,f}-0.5)$              & $1 \cdot dt$    \\\hline
\end{tabular}
\end{table}

The policy is a multilayer perceptron (MLP) with $24 \times 1$ input dimension, $8 \times 1$ output dimension, and two hidden layers of sizes 128 and 64. We use the Exponential Linear Unit (ELU) activation function: 
\begin{equation}
\mathrm{ELU}_\alpha(x)=
\begin{cases}
x, & x \ge 0,\\
\alpha\left(e^{x}-1\right), & x < 0,
\end{cases}
\label{eq:elu}
\end{equation}

The output of the MLP returns the action, an $8 \times 1$ vector containing $\theta_x$ and $\theta_y$ for each of four legs and is interpreted as target angles for the training surrogate’s joints. These are tracked using a fixed PD controller in the simulator. Similar reset handling for time-out and falling is used as in the Anymal locomotion setup \cite{b36}, but we omit termination from stalling to encourage motion.

PhysX training is carried out on an open-loop leg configuration model because closed-loop kinematics are difficult to represent reliably through standard URDF-based pipelines used for PhysX training. The surrogate leg dimensions and action range are chosen to match the intended quadruped leg workspace as closely as possible, with the goal of reducing distortion when transferring the learned policy to the higher-fidelity Simulink inferencing configuration.

\begin{figure}[htbp]
\centering
\includegraphics[width=0.47\textwidth]{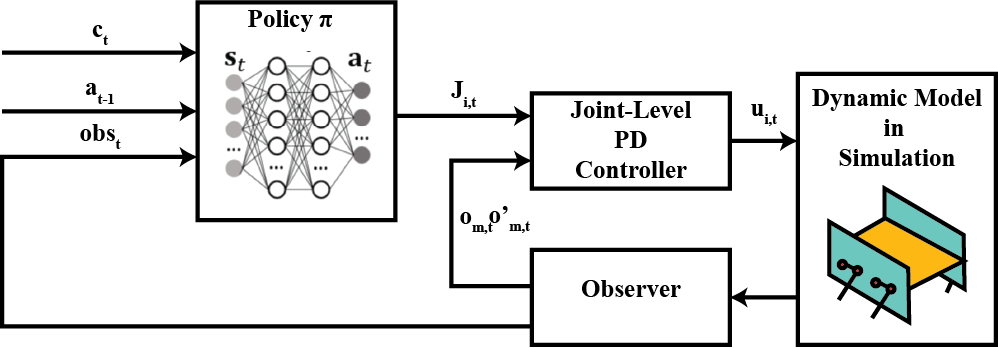}
\caption{Training pipeline: PhysX rollouts in Isaac Gym $\rightarrow$ PPO update $\rightarrow$ compact MLP policy.}
\label{fig:Method3}
\end{figure}


\subsection{Domain Randomization}
To improve robustness to the simulation-to-reality gap in modeling, we apply domain randomization during training. Table~\ref{tab:dr} lists the randomized parameters and distributions. We randomize additive observation/action noise, gravity perturbations, and joint-limit perturbations. We also randomly scale mass, friction, restitution, damping, and stiffness.

\begin{table}[H]
\centering
\caption{Domain randomization parameters used during training.}
\label{tab:dr}
\begin{tabular}{llll}
\hline
Parameter       & Range           & Operation & Distribution \\\hline
Observation     & $[0, 0.002]$    & additive  & Gaussian     \\
Action          & $[0, 0.02]$     & additive  & Gaussian     \\
Gravity         & $[0, 0.4]$      & additive  & Gaussian     \\
Mass            & $[0.05, 0.15]$  & scaling   & Uniform      \\
Friction        & $[0.07, 0.13]$  & scaling   & Uniform      \\
Restitution     & $[0, 0.7]$      & scaling   & Uniform      \\
Damping         & $[0.5, 1.5]$    & scaling   & Uniform      \\
Stiffness       & $[0.5, 1.5]$    & scaling   & Uniform      \\
DoF lower range & $[0, 0.01]$     & additive  & Gaussian     \\
DoF upper range & $[0, 0.01]$     & additive  & Gaussian     \\\hline
\end{tabular}
\end{table}

\subsection{On SC$\mu$M Setup and Inference Simulation}
We evaluate the learned MLP policy in a higher-fidelity closed-loop Simulink Multibody simulation (Fig.~\ref{fig:Method_matlab}) and are able to execute policy inference on SC$\mu$M-3C to study compute-limited update rates. As the “brain” of our full autonomous microrobot, SC$\mu$M-3C is a $2 \times 3 \times 0.3 \space mm^3$ SoC \cite{b10} featuring an ARM Cortex-M0 microprocessor and an IEEE 802.15.4 transceiver \cite{b11}. Importantly, SC$\mu$M-3C also features SRAM and DMEM sizes of 64 kB with a 256 kB SRAM size in more recent versions of the chip, capable of storing our $[128, 64]$ MLP policy. To support hardware-friendly computation, we replace the ELU activation function with Leaky ReLU:
\begin{equation}
\phi(x)=
\begin{cases}
x, & x \ge 0,\\
\alpha x, & x < 0,
\end{cases}
\end{equation}

\begin{figure}[htbp]
\centering
\includegraphics[width=0.5\textwidth]{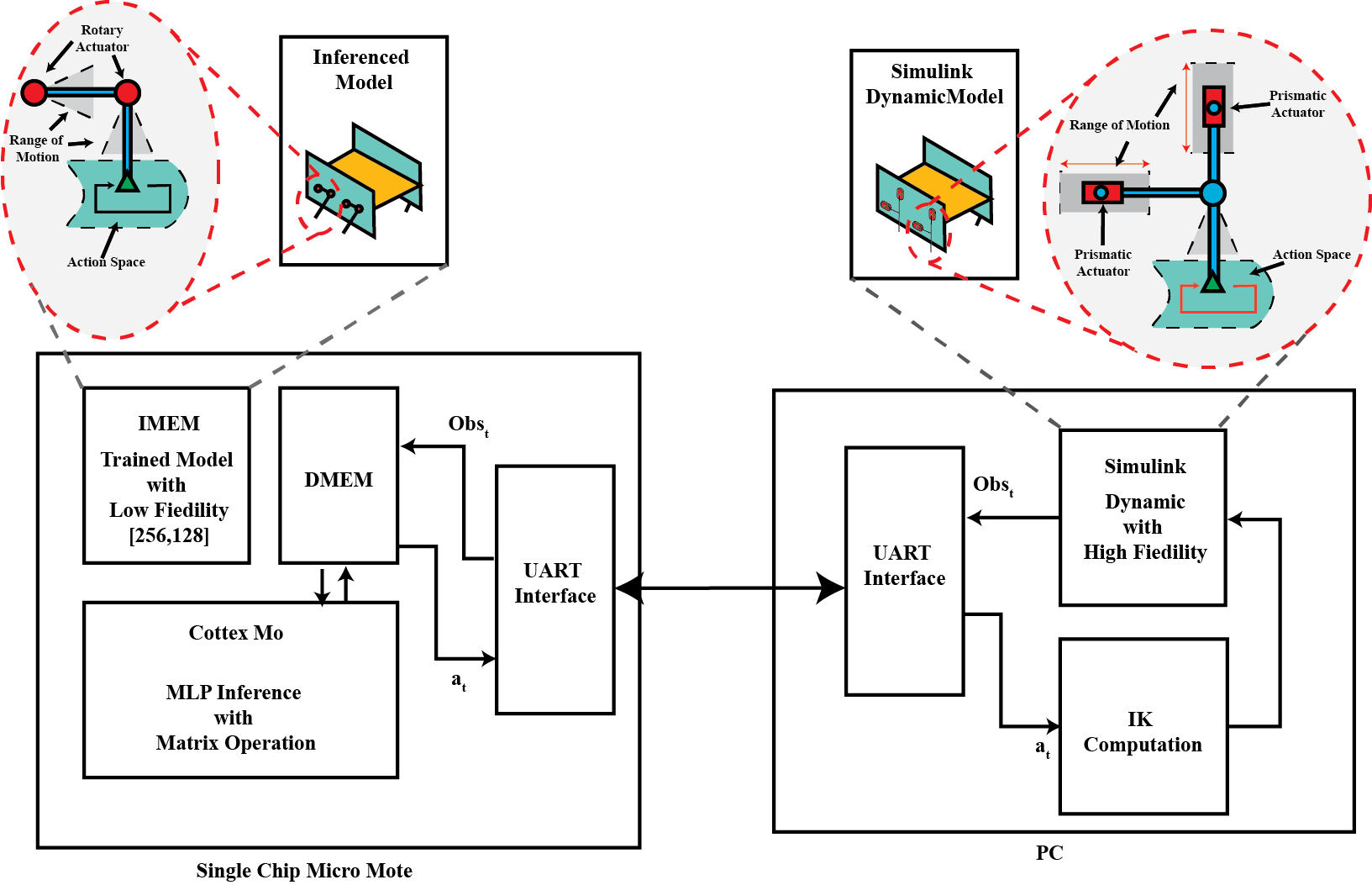}
\caption{Evaluation/inference pipeline: Simulink plant on host PC $\leftrightarrow$ UART $\leftrightarrow$ SC$\mu$M (inference).}
\label{fig:Method4}
\end{figure}

Network inference and simulation tasks are distributed between SC$\mu$M and a host PC as shown in Fig.~\ref{fig:Method4}. The devices communicate over the universal asynchronous receiver-transmitter (UART) protocol. At each step, the host reads the current observation from the Simulink model and takes in a command, then encodes and transmits these to SC$\mu$M. Once the full packet is received, SC$\mu$M runs inference, computing the action. The simulation on the PC pauses while waiting for an action from SC$\mu$M to ensure synchronization. The action is sent back to the PC as the target position for the joints. The joint-level command returned by SC$\mu$M is converted into target positions for the $x$ and $y$ prismatic motors of the closed-loop linkage via inverse-kinematic (IK) equations computed on the PC. Using end-effector targets $(x_{\mathrm{end}},y_{\mathrm{end}})$, motor reference locations, and linkage lengths $(l_x,l_y)$, we use the following equations:
\begin{equation}
\theta_{y} = \sin^{-1}\!\left(\frac{x_{end}-x_{motor,x}}{l_{y}}\right)
\end{equation}
\begin{equation}
\theta_{x} = \sin^{-1}\!\left(\frac{y_{end}+\left(0.5l_{y}\cos(\theta_{y})-y_{motor,x}\right)}{l_{x}}\right)
\end{equation}
\begin{equation}
x_{motor} = x_{end} - 0.5l_{y}\sin(\theta_{y}) - l_{x}\cos(\theta_{x})
\end{equation}
\begin{equation}
y_{motor} = y_{end} + l_{y}\cos(\theta_{y})
\end{equation}

 Once the PC receives the command, it will produce the next timestep observation to send to SC$\mu$M, completing the loop. When analyzing full-loop latency, we mainly consider neural network inference delay since IK computation is relatively lightweight. The Cortex-M0 runs at a 5 MHz clock in our case, constraining the frequency at which we can perform our MLP inference. This frequency corresponds to our update frequency (how fast we can provide updates to our model on PC).

\section{Results}

\subsection{Learned Gait Behaviors and Stability Trends}

Across the observation of gaits for the numerous agents we trained, we qualitatively note a distinct trend of gait selection with respect to command velocity. We estimate that under a command velocity of $\approx 0.025$ m/s, the quadruped will assume a "trot" gait. Between the command velocities of $\approx 0.025$ m/s and $\approx 0.075$ m/s, the quadruped will assume an "intermediate" gait. When command velocity exceeds $\approx 0.075$ m/s, the quadruped will assume a "gallop" gait. We use these three command ranges to group results by gait regime in the remainder of this section.
\vspace{-0.5cm}
\FloatBarrier
\begin{figure}[htbp]
\centering
\includegraphics[width=0.52\textwidth]{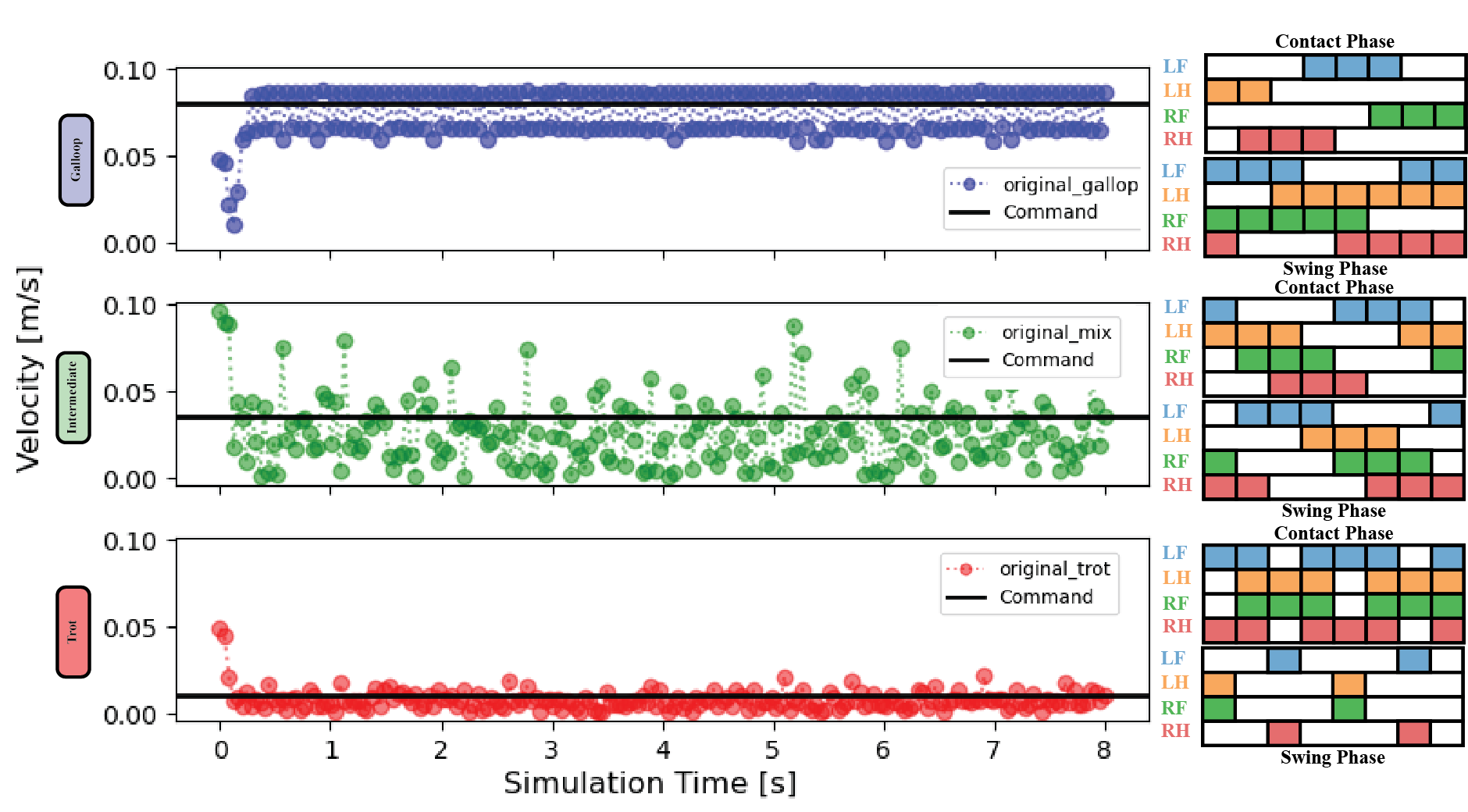}
\caption{FP32 MLP model inference at different command velocities. Blue graph $\rightarrow$ command velocity = 0.08 m/s, "gallop" gait. Green graph $\rightarrow$ command velocity = 0.035 m/s, "intermediate" gait. Red graph $\rightarrow$ command velocity = 0.01 m/s, "trot" gait.}
\label{fig:Result_gait}
\end{figure}

\subsection{Quantization: Per-Tensor vs.\ Per-Feature Int8}
To allow for a higher update frequency on our 5 MHz SC$\mu$M-3C chip, we explore two types of inference speed-up: per-feature Int8 quantization and per-tensor Int8 quantization. To benchmark performance, we observe the gait stability of our unquantized, per-feature, and per-tensor models for different gait velocity commands (0.1 m/s, 0.075 m/s, 0.05 m/s). 
We evaluate Int8 quantization under two common schemes:
\begin{itemize}
    \item \textbf{Per-tensor Int8}: one scale/zero-point per tensor.
    \item \textbf{Per-feature Int8}: separate scales/zero-points per output channel.
\end{itemize}
\begin{center}
\includegraphics[width=0.42\textwidth]{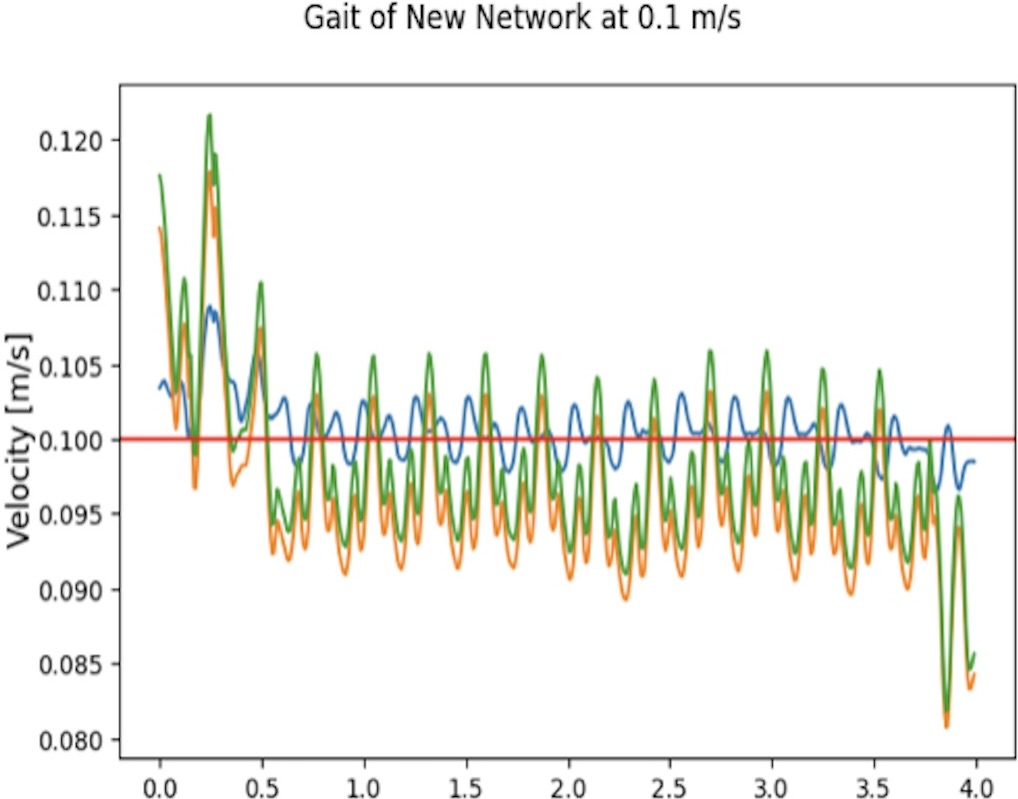}

\vspace{0.5em}
{\footnotesize \textbf{(a)} $v^\star = 0.10~\mathrm{m/s}$}

\vspace{0.9em}
\includegraphics[width=0.42\textwidth]{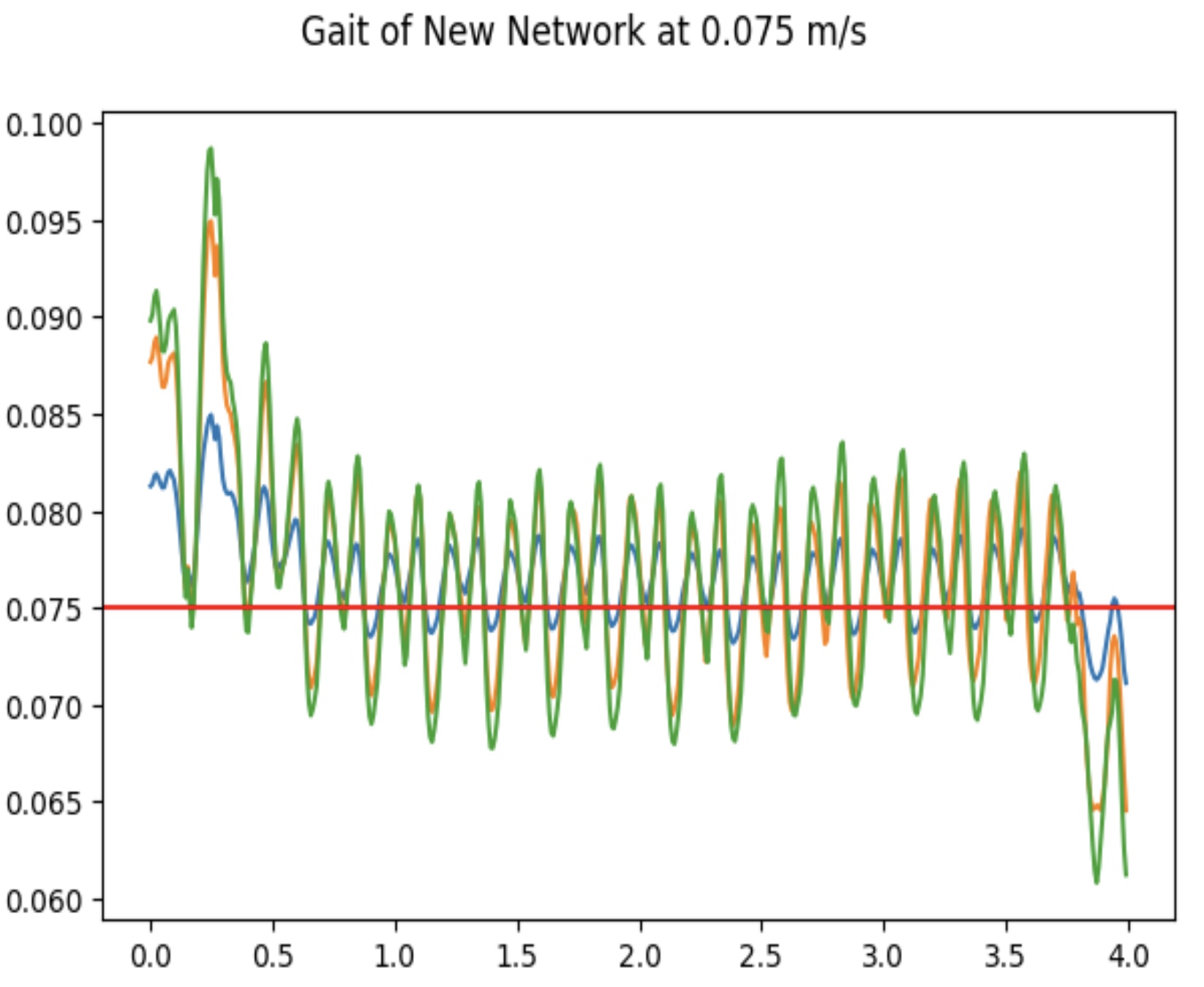}

\vspace{0.5em}
{\footnotesize \textbf{(b)} $v^\star = 0.075~\mathrm{m/s}$}

\vspace{0.9em}
\includegraphics[width=0.42\textwidth]{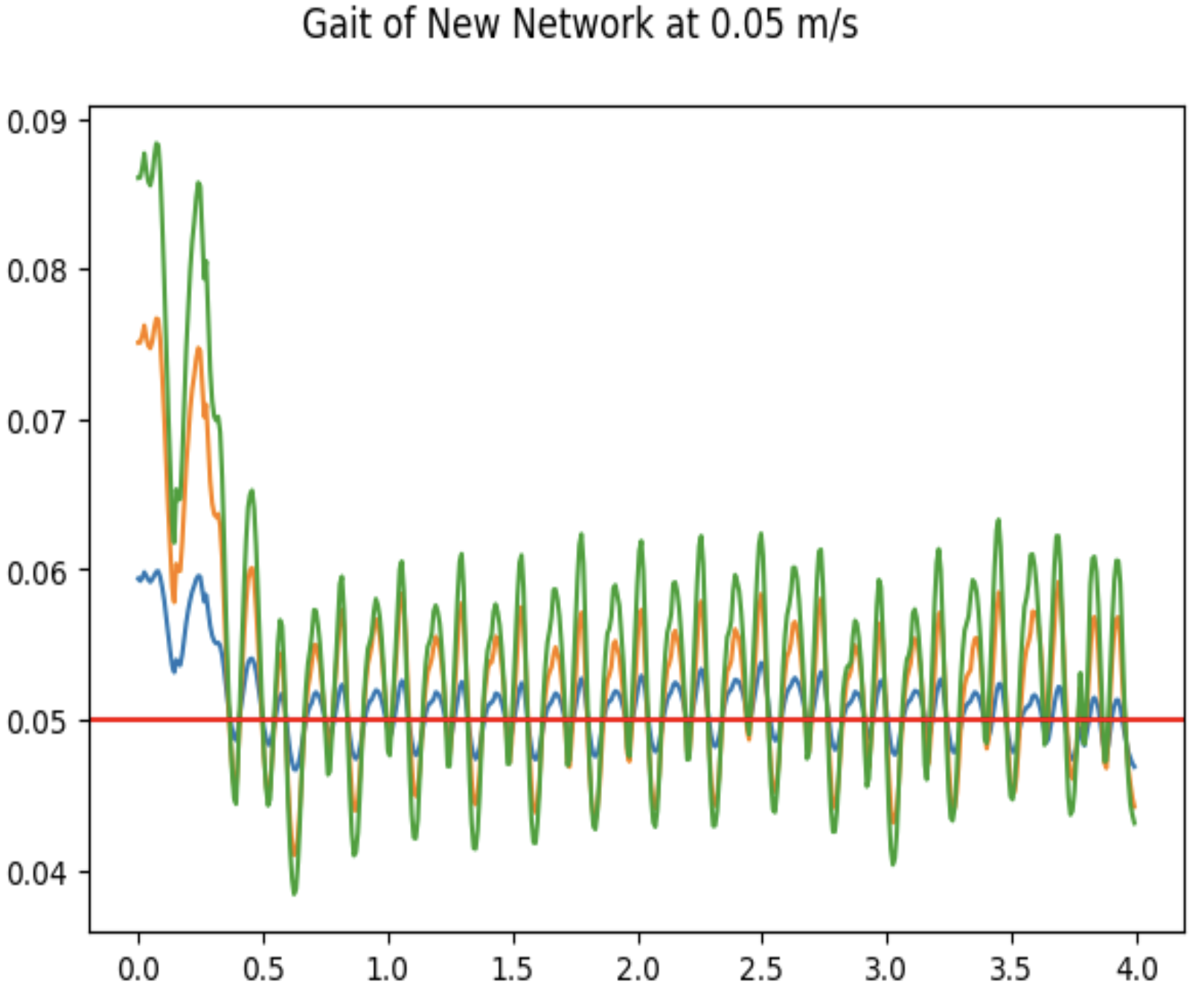}

\vspace{0.5em}
{\footnotesize \textbf{(c)} $v^\star = 0.05~\mathrm{m/s}$}

\captionof{figure}{Microrobot forward velocity tracking vs.\ simulation time for three commanded speeds, comparing FP32 MLP at 120~Hz, per-feature Int8, and per-tensor Int8.}
\label{fig:Result_velocity_track}
\end{center}

Empirically, per-feature Int8 improves gait quality at a given update rate by reducing effective quantization distortion at a small extra compute cost. A representative velocity tracking plot is shown in Fig.~\ref{fig:Result_velocity_track}. We compare Int8 quantization performance on SC$\mu$M against our FP32 MLP at 120 Hz ($f_{update}$ that we trained with). We test at higher command velocities because we observe greater performance degradation for less expressive models on faster gait patterns. Experimentally, we were able to deduce that floating point MLPs on our hardware achieved $\leq 5$ Hz update frequency, resulting in $\approx10 \times$ boost in $f_{update}$ due to quantization.

We compile metrics from our velocity tracking experiment into the table below.
\begin{table}[H]
\centering
\caption{Quantization error and achievable update frequency for different quantization schemes used in Fig. 7}
\label{tab:quant_error_update}
\begin{tabular}{llll}
\hline
\textbf{Quantization Type} &
\textbf{SQNR} &
\textbf{ Error (m/s)} &
\textbf{$f_{update}$ (Hz)} \\\hline
Baseline     & --    & 0.001278  & 120    \\
Per-feature  & 46.57 & 0.004522  & 47.62 \\
Per-tensor   & 29.89 & 0.0048453 & 52.63 \\\hline
\end{tabular}
\end{table}

We select a per-feature Int8 quantized network to balance between inference speed-up and theoretical quantization error. We compare our unquantized and per-feature Int8 neural networks at different gaits with a linear Int8 (1-layer) network as a control.
\begin{figure}[htbp]
\centering
\includegraphics[width=0.5\textwidth]{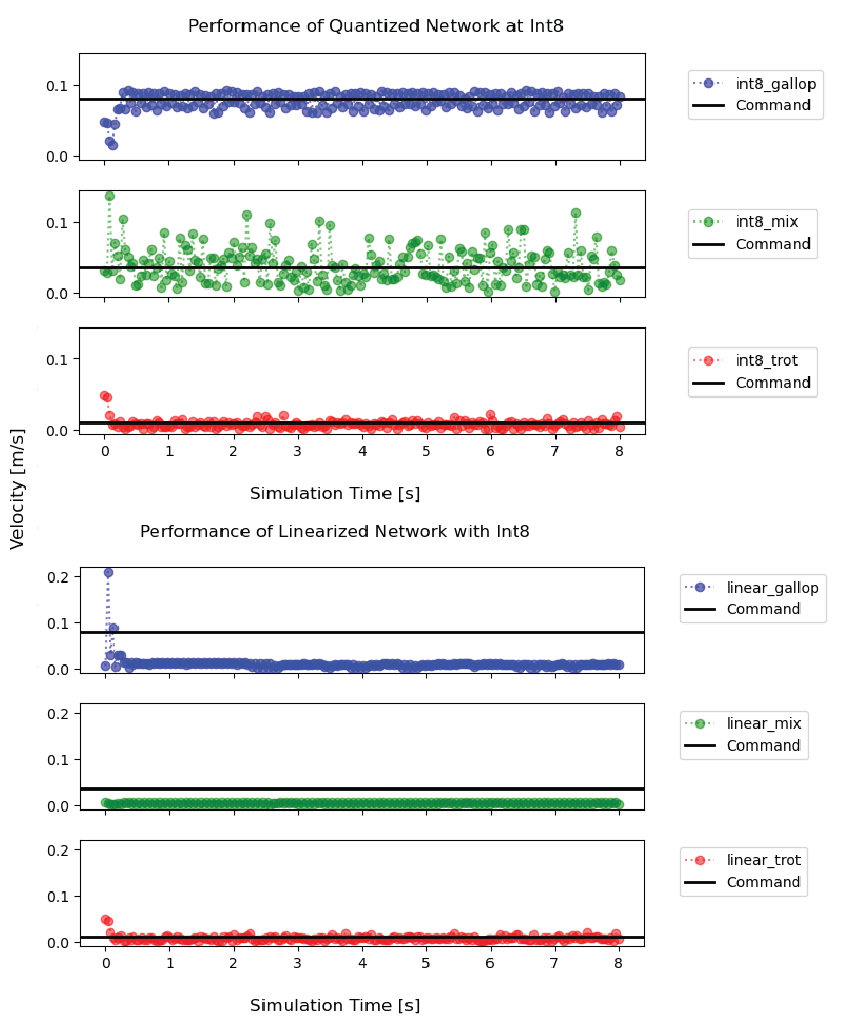}
\caption{Microrobot velocity vs.\ time under command velocities of 0.08 m/s, 0.035 m/s, and 0.01 m/s.}
\label{fig:Result_infer}
\end{figure}

We note differences in model size in the table below. We observe that Int8 quantization is able to achieve acceptable performance on all gait tasks comparable to the unquantized FP32 MLP performance from Fig.~\ref{fig:Result_gait}.

\begin{table}[H]
\centering
\caption{Model size comparison for different network precisions/architectures.}
\label{tab:model_size}
\begin{tabular}{ll}
\hline
\textbf{Network} & \textbf{Size (kB)} \\\hline
FP32 (unquantized) MLP & 204.54 \\
Int8-quantized MLP     & 51.136 \\
Linear baseline        & 1.184  \\\hline
\end{tabular}
\end{table}

\subsection{Update Frequency Sensitivity and Reward Trends}
A core empirical observation is that gait quality depends on inference update frequency: for some gaits (e.g. faster gallop-like behaviors), reduced update frequency causes larger reward degradation than for slower gaits. We summarize this effect using reward ratios (inference reward divided by training reward) as a function of $f_{update}$ across gait regimes.

\begin{figure}[htbp]
\centering
\includegraphics[width=0.47\textwidth]{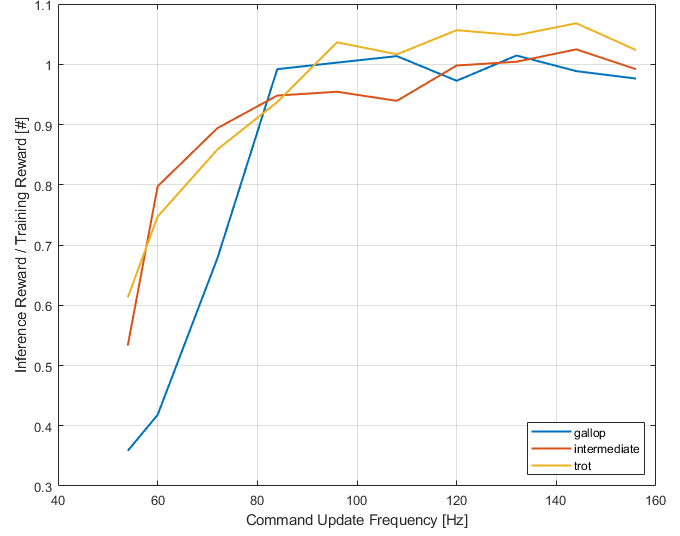}
\caption{Reward ratio vs. command update frequency for multiple gait regimes (trot/intermediate/gallop), averaged over 100 agents.}
\label{fig:Result_rewardfreq}
\end{figure}

\subsection{Real-World Deployment on Macro-Scale Robot}

We were unable to complete the full Sim2Real pipeline for SC$\mu$M-3C because we didn't have a fabricated quadruped microrobot. Instead, we deployed our per-feature Int8 quantized model onto an Arduino with an 8-bit ATmega4809 microcontroller unit and mounted this on a macro-scale quadruped robot. We qualitatively note the acceptable locomotion capabilities of this robot on uneven terrain.

\begin{figure}[htbp]
\centering
\includegraphics[width=0.5\textwidth]{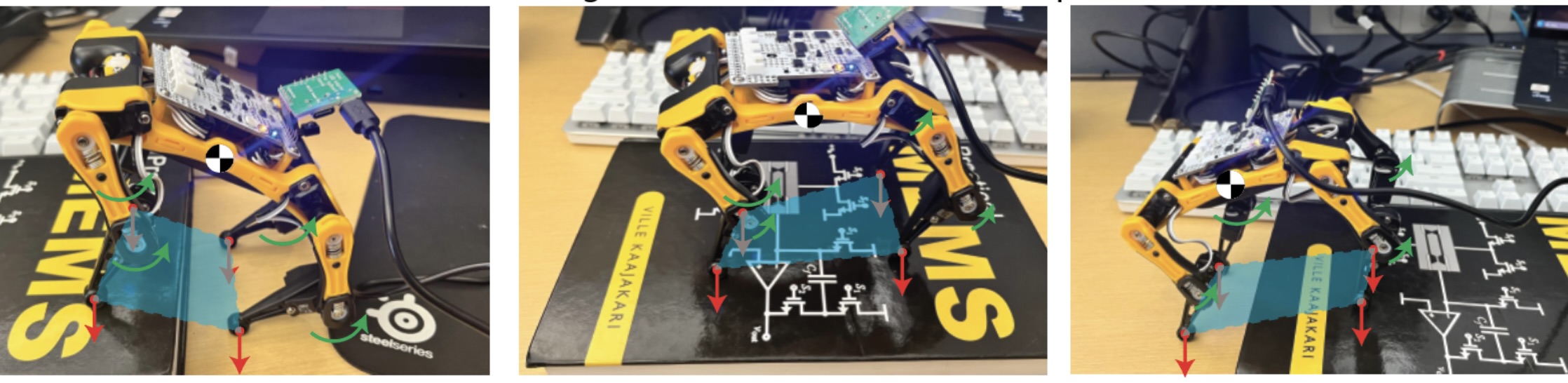}
\caption{Real-world locomotion of macro-scale quadruped on uneven terrain where $f_{update} = 60$ Hz.}
\label{fig:Result_sim2real_big_robot}
\end{figure}
\FloatBarrier
\section{Discussion}

\subsection{Effects of Quantization on Performance}
Our results indicate that update frequency and model parameter quantization-level are critical levers for locomoting quadrupedal, severely resource-limited  microcontroller units. Despite a 4x reduction in model size, our Int8 per-feature policy was able to achieve mm/s-level velocity deviations across all tested microrobot velocity commands with a 10x boost in update frequency. With our closed-loop locomotion, we are able to mitigate our SQNR through quicker updates, keeping our error bounded reasonably. 

\subsection{Real-World Macro-Scale Robot Locomotion}

We qualitatively observed that our macro-scale quadruped was able to maintain forward locomotion over uneven terrain without catastrophic instability. This suggests that the per-feature Int8 quantization for our policy preserves enough
structure to remain functional under real-world disturbances. We reason that our domain-randomized training had improved our tolerance to out-of-distribution perturbations relative to a strictly nominal simulation policy. However, we emphasize that this experiment does not validate true microrobot Sim2Real transfer. Rather, it validates the practicality of the quantized inference and control loop on resource-limited hardware and motivates repeating the same pipeline on the intended microrobot quadruped once fabrication is successful.

\subsection{Accounting for MACs and Activation Functions for the MLP}
In the following sections, we will approximate a model for the number of cycles/update and use this model to discuss gait selection. For a fully-connected layer mapping $n_{\ell-1}\rightarrow n_{\ell}$, the dominant multiply-accumulate
(MAC) count is $n_{\ell-1}n_{\ell}$. Our network has an input and output layer that accepts and outputs a $24\times 1$ and $8\times 1$ vector, respectively,
through two hidden layers $(n_1,n_2)=(128,64)$. For $n_0=24$ and $n_3=8$,
\begin{align}
N_{\text{MAC}}
&= \sum_{l} n_{l-1} n_l \nonumber \\
&= n_0n_1 + n_1n_2 + n_2n_3 \nonumber\\
&= 24\cdot 128 + 128\cdot 64 + 64\cdot 8 \nonumber\\
&= 11776~\text{MACs/update}.
\label{eq:nmac}
\end{align}
Now, we will compute the total parameter count:
\begin{align}
N_{\text{param}} 
&= N_{MAC} + N_{BIAS} \nonumber\\
&= (n_0n_1+n_1n_2+n_2n_3) + (n_1+n_2+n_3) \nonumber\\
&= N_{\text{MAC}} + 200 \nonumber\\
&= 11976~\text{parameters}.
\label{eq:nparam}
\end{align}
Bias operations contribute $n_1+n_2+n_3=200$ additional scalar adds per update, which is small
relative to \eqref{eq:nmac}.

On Cortex-M0/M0+ class cores, inference with Int8 would typically be implemented as scalar
dot-product loops. Int8 weights/activations would be loaded and multiply-accumulated into an
Int32 accumulator. To our knowledge, these cores do not provide SIMD MAC instructions, so
throughput would depend heavily on the specific memory system, loop overhead, and multiplier implementation. Thus, any constant cycles/MAC figure we derive should be interpreted only as an optimistic,
idealized lower bound.

We recognize that non-MAC costs like memory overhead, requantization, and activation functions contribute to overall cycle cost. Therefore, we report the measured end-to-end inference cost
in cycles per update. If SC$\mu$M-3C runs at clock $f_{\text{clk}}$ and achieves an observed update rate $f_{update}$,
then the measured cycles/update are
\begin{equation}
C_{\text{infer}}^{(\text{meas})} \approx \frac{f_{\text{clk}}}{f_{update}}.
\end{equation}

Since this equation captures end-to-end overheads, we use $C_{\text{infer}}^{(\text{meas})}$ directly when mapping clock/power budgets to feasible
update rates. However, it may be useful to decompose $C_{\text{infer}}$ into constituent sources of overhead computation. So far, we have identified the contribution from MAC operations to this variable. Now, we will examine the impact of our activation function on $C_{infer}$. Replacing ELU with the deployed Leaky ReLU function
\begin{equation}
\phi(x)=
\begin{cases}
x, & x \ge 0,\\
\alpha x, & x < 0,
\end{cases}
\label{eq:leakyrelu}
\end{equation}
reduces activation cost and simplifies inference for our resource-constrained hardware. In our MLP,
the number of nonlinear activations evaluated per update is the number of hidden units,
\begin{equation}
N_{\phi}=n_1+n_2=192.
\end{equation}
To concretize our earlier expression for $C_{infer}$ and determine our $f_{update}$ with greater accuracy, we decompose cycles/update as
\begin{equation}
C_{\text{infer}} \approx c_{\text{MAC}}N_{\text{MAC}} + c_{\phi}N_{\phi} + C_{\text{rq}} + C_0,
\label{eq:cinfer_decomp}
\end{equation}
where $c_{MAC}N_{MAC}$ captures cycles needed due to multiply-accumulate operations, $c_{\phi}N_{\phi}$ determines cycles needed for evaluations of our activation function, $C_{\text{rq}}$ finds requantization/clipping costs, and $C_0$ relates to further software/I-O
overhead. Under this equation, changing ELU to LeakyReLU
primarily changes $c_{\phi}$, producing an additive shift in cycles per update.

\subsection{Modeling Per-Tensor vs. Per-Feature Overhead}
Both per-tensor and per-feature quantization share the same dot-product MAC count
$N_{\text{MAC}}$, but differ in how requantization parameters are applied to each layer output.
A generic requantization form can be
\begin{equation}
y_i = \text{clip}_{\text{Int8}}\!\left(\left(\alpha_i \cdot a_i + r\right)\gg s_i\right) + z_i,
\label{eq:requant}
\end{equation}
where $a_i$ is the Int32 accumulator for output neuron $i$, and $\alpha_i$, $s_i$, and $z_i$ are multiplier, shift, and zero-point parameters, respectively. Per-tensor quantization shares parameters across a tensor, while per-feature quantization uses parameters that vary per feature in the output.

The total number of neuron outputs across all layers is
\begin{equation}
N_{\text{neurons}} = n_1+n_2+n_3 = 200.
\label{eq:nouttot}
\end{equation}
We model the cycle costs per update as
\begin{align}
C_{\text{infer}}^{(\text{per-tensor})}
&\approx c_{\text{MAC}}N_{\text{MAC}} + c_q N_{\text{neurons}} + c_{\phi}N_{\phi} + C_0,
\label{eq:cinfer_pt}\\
C_{\text{infer}}^{(\text{per-feature})}
&\approx c_{\text{MAC}}N_{\text{MAC}} + (c_q+c_{\text{load}}) N_{\text{neurons}} + c_{\phi}N_{\phi} + C_0,
\label{eq:cinfer_pf}
\end{align}
where $c_q$ is the requantization compute per output and $c_{\text{load}}$ captures the additional loads per output parameter in per-feature quantization. Although $N_{\text{out,tot}}$ and $N_{\phi}$ are $\ll N_{\text{MAC}}$, the runtime impact still depends on cycles per operation. Requantization and evaluation of activation functions can have much higher per-element costs than a MAC
on Cortex-M0/M0+, so their contributions can be comparable in cycles/update despite smaller counts. However, we anticipate these terms to be dominated, in our specific case, by $c_{MAC}N_{MAC}$.

\subsection{Determining Feasible Update Frequency Based on Power Budget}
We now express total inference cost in cycles per update. If SC$\mu$M-3C runs at clock
$f_{\text{clk}}$ and achieves an observed update rate $f_{update}$, then
\begin{equation}
C_{\text{infer}}^{(\text{meas})} \approx \frac{f_{\text{clk}}}{f_{update}}.
\label{eq:cinfer_meas}
\end{equation}
We use $C_{\text{infer}}^{(\text{meas})}$ directly when mapping hardware budgets to feasible update rates since it includes all overhead. The maximum feasible update rate is
\begin{equation}
f_{update,\max} \approx \frac{f_{\text{clk}}}{C_{\text{infer}}^{(\text{meas})}}.
\label{eq:fumax_meas}
\end{equation}

We can model active power as approximately linear in clock:
\begin{equation}
P_{\text{cpu}}(f_{\text{clk}}) \approx V \cdot I_{\text{MHz}} \cdot \left(\frac{f_{\text{clk}}}{10^6}\right),
\label{eq:power}
\end{equation}
where $I_{\text{MHz}}$ is the active current per MHz at voltage $V$. For a power budget $P_{\max}$, the
maximum sustainable clock is
\begin{equation}
f_{\text{clk,max}} \approx 10^6 \cdot \frac{P_{\max}}{VI_{\text{MHz}}},
\label{eq:fclkmax}
\end{equation}
and the corresponding maximum feasible update frequency becomes
\begin{equation}
f_{update,\max}(P_{\max}) \approx \frac{f_{\text{clk,max}}}{C_{\text{infer}}^{(\text{meas})}}.
\label{eq:fumax}
\end{equation}

\subsection{Resource-Aware Gait Selection}
Let $g\in\{\text{trot},\text{intermediate},\text{gallop}\}$ denote a gait regime. Let $R_g(f_{update})$ denote
the expected reward or reward ratio relative to training reward when executing gait $g$ at inference update frequency $f_{update}$ (Fig.~\ref{fig:Result_rewardfreq}). For a given update frequency $f_{update}$, we select 
\begin{equation}
g^\star=\arg\max_{g}~R_g\!\left(f_{update}\right).
\label{eq:gait_select_fu}
\end{equation}

where $g^\star$ denotes the optimal gait choice. Consequently, given a power budget $P_{\max}$, we determine
$f_{update,\max}(P_{\max})$ from \eqref{eq:fumax} and select
\begin{equation}
g^\star(P_{\max})=\arg\max_{g}~R_g\!\left(f_{update,\max}(P_{\max})\right).
\label{eq:gait_select}
\end{equation}

\subsection{Gait Selection for SC$\mu$M-3C}
For our SC$\mu$M-3C configuration at $f_{\text{clk}}=5$\,MHz, Table~\ref{tab:quant_error_update}
reports $f_{update}\approx 47.62$\,Hz (per-feature Int8). Evaluating
Fig.~\ref{fig:Result_rewardfreq} near $f_{update}\approx 48$\,Hz yields the ordering
$R_{\text{trot}}(f_{update}) > R_{\text{intermediate}}(f_{update}) > R_{\text{gallop}}(f_{update})$. Thus, the trot regime would be selected under this compute/power envelope. Now, we will examine cases where an intermediate or galloping gait is optimal.

Let $f_{update}^{(\text{int})}$ and $f_{update}^{(\text{gal})}$ denote the minimum update frequencies at which the
intermediate and gallop regimes achieve a maximal reward ratio (e.g., $R_g\ge R_{\min}$). From Fig.~\ref{fig:Result_rewardfreq}, $f_{update}^{(\text{int})} \approx 60$ Hz and $f_{update}^{(\text{gal})} \approx 85$ Hz. Using \eqref{eq:fumax_meas}, the clock required to reach a target
update rate $f_{update}^{\text{target}}$ is
\begin{equation}
f_{\text{clk,req}} \approx C_{\text{infer}}^{(\text{meas})}\,f_{update}^{\text{target}}.
\label{eq:fclk_req}
\end{equation}
With an increase in $f_{clk}$ to around 6.25 MHz and 8.85 MHz, we would expect optimality for the intermediate and galloping gaits, respectively. Additionally, we could focus on the reduction of $c_{\{MAC, q, \phi, 0\}}$ through modifications such as optimized algorithms or varying MLP size. This highlights two concrete routes to enabling optimality for different gaits on-device: varying available power/clock, or changing $C_{\text{infer}}^{(\text{meas})}$ via
policy modification or a more optimized kernel.  

As shown in Fig.~\ref{fig:Result_rewardfreq}, controllers exhibiting different gait characteristics perform markedly differently under limited update-frequency constraints. These results motivate the design of dedicated, gait-specific controllers tailored to distinct update-frequency regimes. When the achievable update rate is low, controllers optimized for reduced feedback bandwidth naturally favor statically stable gaits such as trot, whereas higher update rates enable controllers that exploit faster, dynamically demanding gaits such as intermediate or gallop. Rather than relying on a single policy spanning all operating regimes, separate controllers trained for specific ranges of $f_{update}$ can improve both performance and computational efficiency by matching policy complexity, temporal dynamics, and feedback bandwidth to the available compute and power envelope. In addition, for each update-frequency regime, the controller can derive an appropriate command velocity that is consistent with the achievable control bandwidth for the particular gait, thereby avoiding aggressive commands that cannot be reliably executed at lower update rates. This specialization enables deliberate co-design of control policy and hardware operating point, allowing gait and velocity selection to emerge from system-level constraints while minimizing unnecessary computational overhead.

\section{Conclusion}

In this paper, we explored RL-based locomotion for a quadrupedal microrobot under on-device compute constraints. A compact MLP policy trained at 120Hz in massively parallel PhysX-based simulation with domain randomization produced stable locomotion in simulation, developed distinct gait patterns for different ranges of command speeds, and exhibited predictable gait degradation trends under reduced inference update frequency and quantization. We analyzed Int8 quantization (per-tensor, per-feature) and derived cycle/power models for inference on Cortex-M0 class microcontrollers, which led to the formulation of a resource-aware gait selection process that maps device budgets to optimal gait choices. We also deployed our policy onto a macro-scale quadruped and qualitatively observed acceptable locomotion over uneven terrain.

\subsection{Limitations}
This study explores a limited set of hardware-oriented optimizations, focused primarily on Int8 quantization.  We acknowledge that we only determined the reward ratio vs. $f_{update}$ curve for our one trained $[128, 64]$ MLP policy and did not sweep over network size or explore alternative policy architectures. In addition, while we evaluate our policy in a closed-loop Simulink-based pipeline while conducting on-device inference, we do not present a complete Sim2Real experiment on a fabricated quadrupedal microrobot platform. As a result, our proposed gait-selection framework is
validated through simulation trends and calculations rather than real-world physical microrobot trials.

\subsection{Future Work}
Future work includes: (i) full, real-world closed-loop deployment on a fabricated quadruped microrobot, (ii) the exploration of different policy architectures for Sim2Sim or Sim2Real evaluation of quadruped microrobot locomotion, and (iii) extending our gait-selection analysis across policy sizes, architectures, and variations in reward functions.

\section*{Acknowledgment}
We thank members of the Berkeley Sensor \& Actuator Center (BSAC) and collaborators on the SC$\mu$M project. We extend special thanks to Dr. Roberto Calandra for his support.



\end{document}